\documentclass[letterpaper]{article} 
\usepackage{aaai24}  
\usepackage{times}  
\usepackage{helvet}  
\usepackage{courier}  
\usepackage[hyphens]{url}  
\usepackage{graphicx} 
\usepackage{natbib}  
\usepackage{caption} 
\usepackage{algorithm}
\usepackage{algorithmic}
\usepackage{multirow}
\usepackage{array}
\usepackage{hhline}
\usepackage{makecell}
\usepackage{tabularx}
\usepackage[table,dvipsnames]{xcolor}
\newcommand{\red}[1]{\textcolor{red}{#1}}

\usepackage{newfloat}
\usepackage{listings}
\DeclareCaptionStyle{ruled}{labelfont=normalfont,labelsep=colon,strut=off} 
\floatstyle{ruled}
\newfloat{listing}{tb}{lst}{}
\floatname{listing}{Listing}
\title{Small Language Model Can Self-Correct}
\author {
    Haixia Han\textsuperscript{\rm 1},
    Jiaqing Liang\textsuperscript{\rm 2},
    Jie Shi\textsuperscript{\rm 3},
    Qianyu He\textsuperscript{\rm 3},
    Yanghua Xiao\textsuperscript{\rm 1,3}\thanks{Corresponding Author}
}
\affiliations {
    \textsuperscript{\rm 1}Shanghai Institute of AI for Education and School of Computer Science and Technology, East China Normal University\\
    \textsuperscript{\rm 2}School of Data Science, Fudan University\\
    \textsuperscript{\rm 3} Shanghai Key Laboratory of Data Science, School of Computer Science, Fudan University\\
    haixiahan03@gmail.com, \{liangjiaqing, shawyh\}@fudan.edu.cn, \{jshi22, qyhe21\}@m.fudan.edu.cn
}
\usepackage{bibentry}
\begin{document}
\maketitle
\begin{abstract}
Generative Language Models (LMs) such as ChatGPT have exhibited remarkable performance across various downstream tasks. Nevertheless, one of their most prominent drawbacks is generating inaccurate or false information with a confident tone. Previous studies have devised sophisticated pipelines and prompts to induce large LMs to exhibit the capability for self-correction. However, large LMs are explicitly prompted to verify and modify their answers separately rather than completing all steps spontaneously like humans. Moreover, these complex prompts are extremely challenging for small LMs to follow. In this paper, we introduce the \underline{I}ntrinsic \underline{S}elf-\underline{C}orrection (ISC) in generative language models, aiming to correct the initial output of LMs in a self-triggered manner, even for those small LMs with 6 billion parameters. Specifically, we devise a pipeline for constructing self-correction data and propose Partial Answer Masking (PAM), aiming to endow the model with the capability for intrinsic self-correction through fine-tuning. We conduct experiments using LMs with parameters sizes ranging from 6 billion to 13 billion in two tasks, including commonsense reasoning and factual knowledge reasoning. Our experiments demonstrate that the outputs generated using ISC outperform those generated without self-correction. We believe that the output quality of even small LMs can be further improved by empowering them with the ability to intrinsic self-correct.
\end{abstract}

\section{Introduction}
Generative Language Models (LMs) have gained considerable attention due to their remarkable capabilities \cite{guo2023close,suzgun-etal-2023-challenging}. Despite the convincing and realistic nature of text generated by these LMs, a concern with LMs lies in their tendency to produce fabricated facts and generate false information \cite{lin-etal-2022-truthfulqa}. Moreover, these models deliver inaccurate information employing unequivocal expressions, which poses substantial risks as it can lead to the spread of misleading and harmful content.

\begin{figure}[ht]
  \centering
  \includegraphics[width=\linewidth]{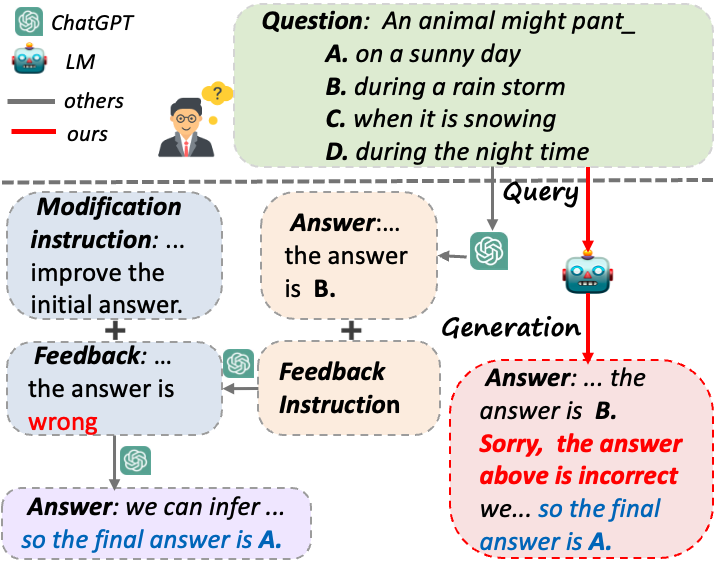}
  \caption{Two self-correction methods are demonstrated in language models in response to a query. The gray line on the left illustrates the process of self-correction employing prompt engineering in large language models like ChatGPT. The red line shows the overall steps of our proposed Intrinsic Self-Correction, where self-verification and self-modification occur spontaneously.}
  \label{Figure1}
\end{figure}

One of the contributing factors to the hallucination lies in inadequate acquisition of knowledge \cite{manakul2023selfcheckgpt,huang2023factual}. For example, consider the question \emph{Which animal is China's nation treasure?}, LMs may provide a different animal name like \emph{tiger} instead of \emph{panda} due to a lack of relevant knowledge. Considerable efforts have been made to alleviate such hallucination induced by lacking of knowledge in LMs. One approach involves supervised fine-tuning LMs with standard ground-truth answers to enhance their comprehension of relevant knowledge \cite{wei2022finetuned,ouyang2022training}. This method has shown promising efficacy. However, it demands a significant amount of high-quality annotated data for training. Additionally, other methods have relied on external verifier or critic model to evaluate the accuracy of a statement \cite{yang-etal-2022-re3,paul2023refiner}. Training a verifier necessitates a large number of high-quality evaluation annotations and further fine-tuning of the model, which restricts its broad applicability to other tasks and domains. 

Another reason for an LM to provide incorrect response is intrinsically linked to the design architecture of generative language models themselves \cite{azaria2023internal,paul2023refiner,shinn2023reflexion}. It is widely acknowledged that LMs generate a sentence by maximizing the likelihood of the next token given all previous tokens. Subtle differences in the preceding sentences can potentially lead to diverse generation outcomes. For example, when the question is \emph{Who is the author of The Analects?}, the model gives the correct answer as \emph{Confucius}. However, when the input question becomes \emph{Is Laozi or Confucius the author of the Analects of Confucius?}, the model is likely to generate an answer of \emph{Laozi}. In this case, the model has the ability to rely on its knowledge to recognize false information \cite{schick2022peer}. This process is akin to how humans perform self-verification of answers to minimize mistakes \cite{flower1981cognitive}. Moreover, when we realize our answer is wrong, we further modify it. Motivated by this, when the model itself detects the potential hallucination, the next step is to correct the error or mistake. Once the model incorporates this inherent self-correction mechanism, it can address similar issues in other domains and achieve self-improvement. 

The existing work \cite{madaan2023selfrefine,ganguli2023capacity} towards self-correction in LMs has mainly focused on lager models like ChatGPT and GPT4, which is challenging to migrate these self-correction methods to small LMs. Some studies indicated that the self-correction ability depends on model parameters and only emerges in models with larger parameter sizes \cite{azaria2023internal}. The main reason is that they devised a sophisticated pipeline and zero-shot prompts to achieve self-correction. However, these prompts crafted for self-verification and self-modification are difficult for small models to understand. As shown in Figure \ref{Figure1}, upon generating the initial answer to the given question, an additional feedback instruction is utilized to guide ChatGPT in generating feedback information regarding the initial answer. This information contains an evaluation of the correctness of the initial answer. A subsequent modification instruction is employed to alter or refine the initial answer based on the feedback received. 

Nevertheless, small models typically lack self-awareness \cite{weng2022large} and tend to exhibit greater confidence in their generated responses. Consequently, they struggle to assess the quality of their generated outcomes. The capability for self-verification serves as a prerequisite for achieving self-correction. Furthermore, the manner in which self-correction is achieved through multi-step prompt engineering within LMs differs from the spontaneous and one-time correction observed in humans.

To empower the capability for self-correction in small language models, we propose \emph{Intrinsic Self-Correction} (ISC), an intrinsic mechanism that relies on two basic abilities: self-verification and self-modification. At its core, the LM provides a response and subsequently evaluates its own answer. Upon identifying an error, the same LM adjusts its initial response. Conversely, if the answer is validated as accurate, no further modifications are required. The self-correction process is not divided into two separate steps, but rather constitutes a single comprehensive step, as depicted by the red arrowed segment in Figure \ref{Figure1}. We trained the LM to process the self-correction through Instruction Fine-Tuning (IFT). For this purpose, we design the data processing procedure to construct the self-correction data and define the data format. During the fine-tuning process, we propose Partial Answer Masking (PAM) to make the model have the ability of self-verification. Our contributions are summarized as follows:
\begin{itemize}
    \item To the best of our knowledge, we are the first to demonstrate that small language models with even 6 billion parameters possess the capacity for self-correction during response generation without relying on ground truth. 
    \item Our proposed \emph{Intrinsic Self-correction} aims to incorporate self-correction as an intrinsic pattern within LM. It involves an independent and spontaneous self-correction process, distinct in nature from existing methods of self-correction that rely on prompt engineering.
    \item To achieve the capability for self-correction in small LMs, we devise a pipeline for constructing self-correction data and define the data format. It can be universally applied to build data for self-correction tasks. Additionally, we introduce a novel training method called  Partial Answer Masking (PAM) to enable the model self-verify its own generated answers.
    \item We conduct experiments on open-source LMs with varying parameter scales to validate the efficacy of our proposed method. The results demonstrate improvements in accuracy across two different datasets.
\end{itemize}
\begin{figure*}[ht]
  \centering
  \includegraphics[width=.8\linewidth]{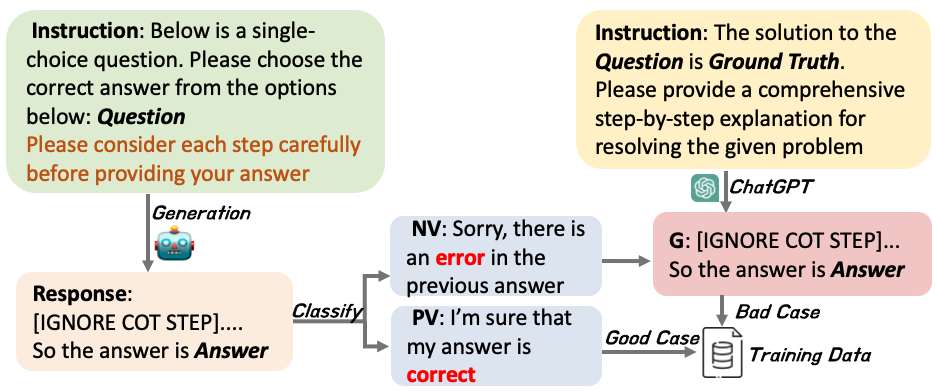}
  \caption{The pipeline of constructing self-correction data.}
  \label{dataConstruct}
\end{figure*}
\section{Related Work}

\textbf{Instruction fine-tuning}
LMs often struggle to maintain alignment well with human intent after pre-training. However, this issue can be alleviated by employing techniques such as Instruction Fine-Tuning (IFT) \cite{stiennon2020learning,ouyang2022training} and Reinforcement Learning from Human or AI Feedback (RLHF/RLAIF) \cite{gao2023scaling,bai2022constitutional}. IFT involves training an LM to generate responses based on provided prompts, which might optionally include task instructions. The data used for fine-tuning is usually less than the original pre-training corpus. These datasets used in IFT are often curated through crowd-sourcing or extracted from an LM that is already capable of generating instruction-following examples. Studies have shown that IFT can enhance zero-shot generalization to various unseen tasks. To optimize the fine-tuning process, various strategies have been proposed to make it more parameter-efficient. These strategies include using adapters \cite{hu2021lora}, prompt tuning \cite{li2021prefix}, and other techniques. 

\textbf{Chain-of thought in language model.} Chain-of-Thought (COT), as an alternative prompt strategy, adopts a chained reasoning approach, incorporating a multi-step reasoning path before generating the final answer. This strategy aims to enhance the model's capacity to handle complex and multi-step queries. \citet{kojima2022large} introduced a novel yet simple approach to prompt the LM by using the phase \emph{Let’s think step by step}, which enables reasoning generation in a zero-shot manner. \citet{zhou2023leasttomost} further decomposed the questions into multiple sub-questions, encouraging the model to engage in a sequential thought process while arriving at the answer. Overall, COT aims to enhance the answer quality through a multi-step reasoning process. 

In contrast, our proposed intrinsic self-correction in LMs involves critiquing and subsequently revising the initially generated results.  It does not fall under the classification of a COT strategy.

\textbf{Self-improvement in language model.} Richer and more detailed feedback plays a crucial role in aligning the output of the LM with the user's preferences and in significantly improving the overall performance of the model. \citet{Welleck2022GeneratingSB} introduced a corrector model used to improve the initial response by learning the specific mistakes made by the LM and the appropriate method for modifying them. Similarly, \citet{paul2023refiner} introduced a critic model to provide feedback on the LM's reasoning errors, evaluating the generated intermediate steps. The feedback, together with the original question and preceding intermediate reasoning steps, was subsequently fed back to the LM to improve the generation of the next step.

Scaling model size can increase model's various capabilities\cite{kaplan2020scaling}, including self-verification (or self-feedback). Recent research efforts have focused on exploring large LMs that employ self-feedback rather than relying on additional models. \citet{madaan2023selfrefine} introduced an iterative self-refinement algorithm that alternates between feedback and refinement. They guided the model to provide feedback about the initial response by employing few-shot prompts. This feedback is then passed back to the same model to help refine the initial response. Some methods used self-improvement to improve the reasoning ability of LM without supervised data. For example, the LM first generates multiple COT reasoning paths and corresponding answers for each question \cite{wei2022chain}. The answer with the highest consistency is selected by majority voting. The  COT reasoning paths exhibiting high-confidence, along with the selected answers, are augmented by mixed formats to serve as the training data used for IFT. Additionally, the LM verified its multiple generated answers in turn and calculated the verification score based on the number of predicted masked values \cite{weng2022large}. Instead of requesting the LM to answer the same query multiple times, a more effective approach is to prompt the LM to rephrase the query and then answer each rephrased question individually. 

We also focus on providing self-feedback and enabling self-modification to enhance the quality of generation of LMs. Furthermore, we extend this capability to encompass more generalized models, including smaller LMs. 
\begin{table*}
\caption{Training Data samples}
\begin{tabular}{l}
     \hline
      \textbf{Question:} \red{\emph{(TaskP)}} Examine the following options carefully and select the correct one. \red{\emph{(COTP)}} Before providing your final \\answer, give the analysis steps. \red{\emph{(SCP)}} And you need double-check your response for accuracy before proceeding to submit.\\
\red{\emph{(question)}} Where do you buy tickets at a ticket booth for games?\\
A. train station B. cathedral C. metro station D. fairgrounds E. amusement park \\

\textbf{Answer:} \red{\emph{($A_{1}^{1}$- COT)}} The question mentions the keywords ``buying tickets" and ``games", so we can guess that this is a $\cdots$ \\
\red{\emph{($A_{1}^{1}$)}} Therefore, the correct answer is \textbf{D. amusement park} \\
\emph{(PV)} Thinking about the correctness of the previous answer$\cdots$ \\
Thinking result: I am sure that the earlier answer is \textbf{correct} \\
\hline
\textbf{Question:} \red{\emph{(TaskP)}} Please choose the most appropriate one from the following options:\\
\red{\emph{(question)}} what contributes more, though less than some believe, to the lung damage caused by smoking?\\
A. smoking less B. switching to chewing C. no filters D. switching to e-cigs\\
\red{\emph{(COTP)}} Please give the detailed solving process and \red{\emph{(SCP)}} verify your response before final submission. \\
\textbf{Answer:} \red{\emph{($A_{2}^{1} $-COT)}}\colorbox{GreenYellow}{ Smoking causes less lung damage than people think, but it's not completely without effect.} \\
\colorbox{GreenYellow}{So the answer is \textbf{A. smoking less}}\\
\emph{\red{(NV)}} Thinking about the correctness of the previous answer $\cdots$ \\
Thinking result: Sorry, there is an error in the previous answer. 
\\
\red{\emph{(SC-COT)}} Let's analyze each option: \\
A. smoking less: The question clearly mentions that it contributes less than some people think,$\cdots$ $\cdots$\\
C. no filters: filters  it contributes to lung damage, and to a lesser extent than some believe. Therefore, the\\
no filter option meets the requirement. \\
\red{\emph{($A_{2}^{2}$)}} So the correct option is \textbf{C. no filter}.
\\
\hline
\end{tabular}
\label{dataFormat}
\end{table*}

\section{Methods}
\subsection{Task Formulation}
In this paper, Intrinsic Self-Correction is employed to accomplish auto-regressive tasks. Given an input sequence $x$, an LM $M$ is tasked with generating an output $y$. Typically, to generate a correct or plausible output, the model needs to incorporate explanation or reasoning, denoted as $z$, as intermediate steps. The process can be described as follows: 
\begin{equation}
p(z,y_0|x) = p(y_0|x,z)p(z|x), 
\end{equation}
where $y_0$ denotes initial output. The same model $M$ provides self-verification of the output to assess the accuracy of $y_0$, represented as $p(v | x,z,y_{0})$, where $v$ represents the verification of $y_0$ and it is a binary feedback signal. Upon detecting a fault, the LM $M$ proceeds to self-modify and generate a revised answer $y$. If no errors are found, there is no need for any modification, and in such case, $y = y_0$. Ultimately, $y$ represents the final response of the model to the input sequence, as given by:
\begin{equation}
    p(y|x)=p(y|x,z,y_0,v)
\end{equation}

It is important to note that the stages of generating initial output, self-verification, and self-modification (if necessary) are not carried out separately. Rather, they are accomplished with just an instruction, referred to as the Self-Correction Prompt (SCP). There are some prompt examples of SCP: \emph{double-check your response for accuracy before proceeding to submit}, \emph{before you finalize your answer, please reexamine it to ensure its correctness} and \emph{please review your response carefully to make sure it is correct before submitting}. 

\subsection{Self-correction Data}
We mimic human self-correction behavioral patterns to design self-correction data format as shown in Table \ref{dataFormat}. We define the self-correction data format as a Question-Answer pair. Next, we will elaborate on how to construct each component separately.

\textbf{Question preparation. }We utilize an LM $M$, which is capable of following instructions, to generate answers for a set of questions. These questions are randomly sampled from datasets on various tasks and come with ground truth.

To ensure the quality of the generated answers, we leverage the COT to instruct the model $M$ to initially generate a problem-solving process and finally provide an answer to the given question. Therefore, we design a range of diverse COT prompts (COTPs) to guide the $M$ to generate the COT analysis. For example, we use prompts like, \emph{Please select the correct option from the provided choices and offer a comprehensive problem-solving process}. In the few shot setting, the whole prompt also contains other question-COTPs and answer examples. We combine task prompts (TaskP) for responding to specific question types, instructions for producing COT analysis processes (COTP), instructions for self-correction (SCP), and the question itself. This constitutes the \emph{Question} part for the self-correcting data.

Additionally, we use two methods to enhance the model's understanding of various self-correction instructions. First, we use gpt-3.5-turbo to generate diverse prompts for different types of instructions mentioned above. We present a brief prompt template below \emph{I want you act as a Prompt Rewriter. Your objective is to rewrite a given prompt to make language model to understand. The rewritten prompt must ensure that the requirement remains unchanged. \#Given Prompt\#:$\left[TaskP \mid COTP \mid SCP \right]$}. Second, we randomly select one instruction from TaskP, COTP and SCP sets and combine them. The order of these instructions can also be rearranged, but TaskP is positioned at the beginning.

\begin{table*}
\begin{tabular}{cccccc}
     \hline
     \multirow{2}{*}{\textbf{Base Models}}&\multicolumn{2}{c}{\textbf{OpenBookQA}} & &\multicolumn{2}{c}{\textbf{CommonsenseQA}} \\
     \cline{2-3} \cline{5-6}
     & \textbf{ACC-First} &\textbf{ACC} & &\textbf{ACC-First} &\textbf{ACC}\\
     \hline
     CuteGPT-7B &25.2 &29.0 (+3.8) & &23.2 &28.9 (+3.7) \\
     CuteGPT-13B &37.2 &42.0 (+4.8) & &35.6 &37.9 (+2.3)\\
     Llama2-7B &52.2 &52.2 (+0.0) & &52.2 &52.3 (+0.1) \\
     ChatGLM-6B &37.0 &42.6 (\textbf{+5.6}) & & 34.3 &38.7 (\textbf{+4.4}) \\
     Vicuna-7B &28.6 &28.80 (+0.4) & &25.9 &26.2 (+0.3) \\
     Vicuna-13B &33.8 &34.0 (+0.2) & &32.4 &32.6 (+0.3) \\
     \hline
\end{tabular}
\centering
\caption{Main results of several open-source LMs. After self-correction, their accuracy is improved on test datasets.}
\label{baseline}
\end{table*}

\textbf{Answer preparation.} To ensure answer diversity, we utilize nucleus sampling \cite{holtzman2019curious} to generate multiple answers. After generating multiple answers for a question, the next step is to evaluate the accuracy of each answer by comparing it with the provided ground truth for the each question. In the case of multiple-choice questions, we extract the options of the final answer through string matching and then directly compare them with the standard answer to check the accuracy.

For a good case, the outcome of self-verification should be positive, indicating there is no need to modify the initial answers. Accordingly, given the question $x$, the \textbf{Answer} is represented as ($A_{1}^{1}$-\emph{COT}$||A_{1}^{1}||PV $), where $A_{n}^{i}$ represents the model attempts $n$ times to obtain the correct answer, and the current answer is the $i$th response, $A_{n}^{i}$-COT represents the COT process of the answer, $PV$ denotes the positive verification in self-correction, and  $||$ represents the concatenation. Here, we set the verification as a binary signal. A positive verification can be set like \emph{I am sure my answer is correct.}

 Conversely, for a bad case, negative self-verification result is excepted, such as \emph{Sorry, there is an error in the previous answer}. It requires modifications to the initial answer. To enhance the model's ability to generate more appropriate reasoning process and correct answer, we utilize the ground truth, representing the standard answer, as the revised answer. Additionally, we employ gpt-3.5-turbo to assist in generating the COT analysis process, denoted as $G$. We use prompts like \emph{the answer of [Question] is [Ground Truth]. Please provide a step-by-step explanation for resolving the given problem}. Therefore, the data format is ($A_n^{1}$-COT$ ||A_n^{1}||NV||A_n^{2}$-COT$||A_n^{2}\cdots A_{n}^{n}||PV)$, where $NV$ indicates negative verification. In Table \ref{dataFormat}, we provide two general examples of self-correction data, representing examples where the correct answer is obtained without correction and with one correction respectively. We also provide detailed prompt examples used at each step in the Appendix.

This pipeline can be utilized to customize the self-correction data for various corrections, depending on the specific task type. The general process of constructing self-correction data is shown in Figure \ref{dataConstruct}. 

\subsection{Partial Answer Masking}
In the instruction fine-tuning stage, the goal is to guide the model to follow human intention by referring to annotated data. Usually, only the loss associated with the answer component of the training data is used for gradient back-propagation, while the loss related to the input is not employed in updating weights. 

In our paper, we apply this loss calculation method to good cases. However, it is not suitable for bad cases due to two reasons. First, these bad cases contain error information, which increases the likelihood of the common issue of hallucination inherent in language models. Second, deliberately training the model to first generate incorrect answers and then correcting them does not align with our intention of teaching the LM how to self-correct. Our aim is instead to enable the model to spontaneously correct itself when it generates information inconsistent with its internal knowledge. Therefore, for bad cases, we refrain from computing the loss corresponding to the incorrect answer part in the output during training. Instead, we calculate the loss from the output on self-verification. It means that for bad cases, only the self-verification and modified correct answer contribute to the loss calculation and parameter update. We call this training method Partial Answer Masking (PAM). In Table \ref{dataFormat}, the underline part in the output is excluded from the loss calculation.  

\section{Experiments}
\subsection{Experimental Settings}

\textbf{Datasets.} We conduct experiments on two question-answering datasets, including OpenBookQA$\footnote{\url{http://data.allenai.org/OpenBookQA}}$ and CommonsenseQA$\footnote{\url{https://www.tau-nlp.sites.tau.ac.il/commonsenseqa}}$. OpenBookQA is a science question-answering dataset, containing 5,957 elementary-level science multiple-choice questions with 4 options each. These questions evaluate human comprehension of 1,326 core science facts and their application to novel scenarios. CommonsenseQA is a single choice question-answering dataset that necessitates diverse forms of commonsense knowledge for accurate answer prediction. It comprises 12,102 questions with 5 choices each. After performing our proposed self-correction data construction process on the two datasets, the training data comprises about 15,000 self-correction samples. We use another about 1,700 examples as test data, of which 500 are from OpenBookQA and 1,200 are from CommonsenseQA.\\

\begin{table*}
\begin{tabular}{ccccccc}
     \hline
     \textbf{Base Models} &\textbf{Confidence} &\textbf{EvalACC} &\textbf{R2R} &\textbf{R2W} &\textbf{W2W} &\textbf{W2R}\\
     \hline
\rowcolor{gray!10}
    \multicolumn{7}{c}{\emph{\textbf{OpenBookQA}}} \\
     
     CuteGPT-7B &70.6 &36.6 &9 &20 &78 &39\\
     CuteGPT-13B &40.8 &52.0 &57 &54 &57 &78 \\
     Llama2-7B &99.6 &\textbf{52.2} &0 &0 &2 &0\\
     ChatGLM-6B &60.4 &27.2 &34 &31 &76 &59 \\
     Vicuna-7B &98.6 &28.8 &0 &0 &6 &1 \\
     Vicuna-13B &97.2 &34.2 &1 &4 &4 &5\\
     \hline
  \rowcolor{gray!10}
    \multicolumn{7}{c}{\emph{\textbf{CommonsenseQA}}}\\
  
     CuteGPT-7B &87.7 &26.8 &10 &6 &83 &51\\
     CuteGPT-13B &42.2 &\textbf{53.8} &111 &131 &111 &159 \\
     Llama2-7B &99.8 &52.3 &0 &0 &4 &2\\
     ChatGLM-6B &52.7 &52.3 &70 &92 &272&146\\
     Vicuna-7B &98.9 &26.0 &1 &2 &13 &9\\
     Vicuna-13B &97.5 &33.1 &5 &5 &13 &8\\
     \hline
\end{tabular}
\centering
\caption{The quantitative results of Intrinsic Self-Correction on two tasks.}
\label{QS}
\end{table*}

\textbf{Base language models.} Our goal is to evaluate whether we can improve the performance of any LMs through our proposed ISC, even to small LMs. We opt for open-source models with parameters ranging between 6 billion and 13 billion. We use CuteGPT-7B, CuteGPT-13B$\footnote{\url{https://github.com/Abbey4799/CuteGPT/}}$, ChatGLM-6B$\footnote{\url{https://github.com/THUDM/ChatGLM-6B}}$, Llama2-7B\footnote{\url{https://huggingface.co/meta-llama/Llama-2-7b-hf}}, Vicuna-7B\footnote{\url{https://huggingface.co/lmsys/vicuna-7b-v1.3}} and Vicuna-13B\footnote{\url{https://huggingface.co/lmsys/vicuna-13b-v1.3}} as our instruction fine-tuning base models for continuing training. These models have been fine-tuned on the instruction data and have the ability to follow instructions. CuteGPT is based on the original Llama model structure, expands the Chinese vocabulary and performs pre-training. CuteGPT has two public versions: CuteGPT-7B and CuteGPT-13B. ChatGLM-6B also is an open bilingual language model. It is trained for about 1T tokens of Chinese and English corpus, and demonstrates outstanding performance among language models of equal parameter size. Llama2 is a family of state-of-the-art open-access large language models released by Meta, ranging in scale from 7 billion to 70 billion parameters. Vicuna is also an open-source chatbot trained by fine-tuning Llama.

In the instruction fine-tuning stage, these models employ distinct training strategies. CuteGPT family models employ full fine-tuning, Llama2-7B and Vicuna family models utilize Low-Rank Adaptation (LORA), and ChatGLM-6B employs Prompt-tuning.

\textbf{Metrics. }In our experiment, a model is given two chances to answer questions. It only attempts to correct itself after making an error in the first response. To evaluate the accuracy of generation answer, we employ string matching to extract the LM's option answers and compare them with the ground truth provided in datasets. We use several evaluation metrics to assess the various performance of the ISC. We introduce the following evaluation metrics:\\
\begin{itemize}
    \item ACC-First: it represents the accuracy of the initial answer of LM.
    \item ACC: it denotes the accuracy after correction of LM.
    \item EvalACC: this metric represents the probability of the LM correctly evaluating its own initial answer. It reflects the self-verification ability of the LM.
    \item Confidence: it refers to the level of confidence of LM, which represents the proportion of questions where LMs assess their answers as correct.
\end{itemize}

Besides, in our experiments, to assess the capability for self-modification of LM, we use R2R, R2W, W2R, and W2W to denote the times that the LM modifies the correct answer to the right answer, changes the right answer to the wrong answer, changes the wrong answer to another incorrect answer, and modifies the wrong answer to the right answer, respectively.
\begin{table*}
\resizebox{\linewidth}{!}{
\begin{tabular}{cccccccccc}
     \hline
     \multirow{2}{*}{\textbf{Base Models}}&\multicolumn{4}{c}{\textbf{OpenBookQA}} & &\multicolumn{4}{c}{\textbf{CommonsenseQA}}\\
     \cline{2-5} \cline{7-10}
     &\textbf{ACC-First} &\textbf{ACC} &\textbf{Confidence} &\textbf{EvalACC} & & \textbf{ACC-First} &\textbf{ACC} &\textbf{Confidence} &\textbf{EvalACC} \\
     \hline
     CuteGPT-7B w/ PAM &25.2 &29.0 &70.6 &36.6 &  &23.2 &26.9 &87.7 &26.8\\
     CuteGPT-7B w/o PAM &26.0 &28.2 &2.0 &68.6 & &18.8 &31.2 &2.2 &76.2\\
     \hline
     CuteGPT-13B w/ PAM &37.2 &42.0 &40.8 &52.0 & &36.6 &37.9 &42.2 &53.2  \\
     CuteGPT-13B w/o PAM &28.6 &32.4 &1.6 &68.2 & &20.5 &34.2 &1.9 &77.8\\
    \hline
     ChatGLM-6B w/ PAM &37.0 &42.6 &60.4 &27.2 & &34.3 &38.7 &52.7 &52.3\\
     ChatGLM-6B w/o PLM &30.2 &36.0 &56.6 &32.2 & &26.0 &32.9 &50.0 &47.2 \\
     \hline
     Vicuna-7B w/ PAM &28.6 &28.8 &98.6 &28.8 & &25.9 &26.2 &98.9 &26.0 \\
     Vicuna-7B w/o PAM &23.4 &23.6 &97.4 &25.0 & &13.8 &13.9 &39.8 &14.0\\
     \hline
     Vicuna-13B w/ PAM &33.8 &34.0 &97.2 &34.2 & &32.4 &32.6 &97.5 &33.9\\
     Vicuna-13B w/o PAM &32.0 &31.2 &89.6 &32.6 & &31.5 &32.4 &97.5 &33.1\\
     \hline
\end{tabular}}
\caption{The impact of Partial Answer Masking on capability for self-correction.}
\label{ablation}
\end{table*}

\subsection{Results and Analysis}

\textbf{Main Results.} We conduct experiments to analyze the accuracy after applying ISC on the test data. The results are presented in Table \ref{baseline}. Following the application of self-correction, a noticeable enhancement in accuracy is observed across all models for the two given tasks. This implies that the integration of ISC provides the base models with self-correction capabilities. This intrinsic ability allows the model to modify the answer when it detects an error in the initial response generation. For example, in the case of ChatGLM-6B, correcting answers yields a notable improvement of 5.6\% in accuracy on the OpenBookQA dataset, improving it from 37\% to 42.6\%.

The effectiveness of enhancing the self-correction of small LMs depends on the innate abilities. The variations in effects of enhancement differ among models with similar parameter scales. For instance, among these base models, ChatGLM-6B features the smallest parameter scale, yet it demonstrates the most prominent improvement. On the other hand, while Llama2-7B attains the highest accuracy in its answers, making corrections to its responses proves to be challenging.

\textbf{Quantitative Analysis.} In the part, we delve further into the analysis of the intrinsic self-correction of base models. The results are shown in Table \ref{QS}. 

\emph{Discovery 1:} When base models lack strong inherent capabilities but exhibit a high degree of confidence in their generated outcomes, the accuracy of self-verification tends to be notably lower. The performance gains derived from self-correction remain constrained. Taking Vicuna-13B as an example, it shows high confidence in its responses, which makes it challenging to accurately evaluate its initially generated answer, resulting in minimal attempts to modify its initial responses.

\emph{Discovery 2:} Observing the values in the W2R column, we find that if an LM can recognize errors in its responses and attempt to modify the initial answers, an opportunity arises to transform them into accurate solutions. This supports the earlier assumption that hallucination in LM output is not solely attributed to knowledge deficits, but is also linked to contextual texts. In this case, relying on the model itself for correction becomes feasible.

\emph{Discovery 3:} The values in the W2W column also constitute a significant portion of the total modifications. However, despite undergoing self-correction, the models have not achieved successful revisions. Two factors account for this phenomenon. Firstly, a single corrective step is not enough to help LM answer correctly, potentially leading to persistent incorrect answers. Employing multiple iterations of corrections could potentially reduce the W2W values to some extent. Secondly, the persistence of incorrect responses by the model could primarily result from knowledge insufficiency. The infusion of relevant domain knowledge is a method to be considered.

\emph{Discovery 4:} By examining the values within the R2R column, we discern that even though our experiments primarily focused on self-verification of the final answers rather than the problem-solving process, we note that the models, through self-correction, could identify inadequacies in the analytical process and subsequently provide supplementary analysis.

We conduct case study and present several instances about W2R and R2R cases in the Appendix.
\begin{figure}[t]
  \centering
\includegraphics[width=\linewidth]{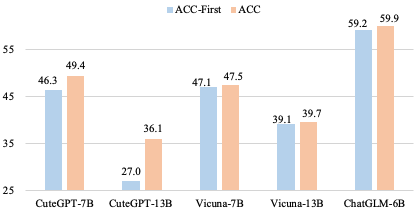}
  \caption{Zero-shot performance of Intrinsic Self-Correction on the StrategyQA test data. After the second round of correction, the accuracy improves obviously.}
  \label{strategy}
\end{figure}

\textbf{Ablation Analysis.} In this part, we verify the impact of the PAM on the LM's self-correction through ablation analysis. Unlike PAM, a commonly used training method in IFT involves utilizing the entire answer for loss calculation. By employing identical data, model architecture, and hyper parameters settings, we conducted a comparative evaluation of the effects of the PAM on self-correction, as illustrated in Table \ref{ablation}. 

Except for CuteGPT-7B without PAM, which exhibits a slightly higher initial answer accuracy on the OpenBook dataset compared to CuteGPT-7B with PAM, nearly all results indicate that employing the PAM leads to higher accuracy in generating answers. Furthermore, the improvement in answer quality through self-correction is most prominent after utilizing the PAM.

Training without PAM appears to reduce the accuracy of self-verification, which is particularly evident in the CuteGPT family models. This phenomenon arises from an imbalance in our constructed training data, where the number of bad cases outweighs that of good cases. However, such data imbalances do not impact the results when employing PAM. This is because, in IFT with PAM, the model effectively learns from the correct information of answer while disregarding the incorrect information. Without PAM, the model learns from the entire answer, including incorrect  information as well, leading the model to evaluate answers as incorrect in most cases. The relatively low accuracy of an LM, assessing answers as incorrect helps enhance the accuracy of self-verification. Additionally, for the Vicuna family models, regardless of the training method employed, they consistently exhibit strong confidence in their generated outputs, making them refuse to self-correct.

\textbf{Zero-shot Performance on New Task. }We evaluate the generalization ability of using ISC on a novel task. We choose StrategyQA$\footnote{\url{https://github.com/eladsegal/strategyqa}}$  as the new test task, which serves as a question answering benchmark focusing open-domain inquiries. This task requires providing either  ``true'' or ``false'' as a response to the given questions. The results are presented in Figure \ref{strategy}.
 
We discover that ISC remains effective for the new task. After the second round of correction, the accuracy of all LMs improve. 

\section{Conclusion}
We introduce Intrinsic Self-Correction (ISC) in LMs, an approach that utilizes models' own capabilities to identify and further modify their initial responses autonomously. This strong capability can even be applied to smaller LMs. We first devise a general process for constructing self-correction data,  applicable to generating diverse self-correction task training data. Furthermore, we introduce a novel fine-tuning method named PAM to instruct LMs to self-correct. We conduct experiments on several open-source LMs to validate the efficacy of ISC. The experimental results on two distinct tasks consistently demonstrate that the utilization of ISC empowers the models with the capability for self-correction, and improves the accuracy of generated answers. In the best case, the accuracy enhancement reaches up to 5.6\%. We also conduct a comprehensive analysis of the ability of the base model to self-validate and self-correct after using ISC. These findings help us understand how ISC works better. 
\section{Acknowledgements}
Yanghua Xiao is also a member of Research Group of Computational and AI Communication at Institute for Global Communications and Integrated Media, Fudan University.

This work was supported by Science and Technology Commission of Shanghai Municipality Grant (No. 22511105902), Shanghai Municipal Science and Technology Major Project (No.2021SHZDZX0103), and National Natural Science Foundation of China (Grant No.62102095).
\bibliography{Latex/isc}
\end{document}